\documentclass[10pt,twocolumn,letterpaper]{article}

\usepackage{cvpr}
\usepackage{times}
\usepackage{epsfig}
\usepackage{graphicx}
\usepackage{subcaption}
\usepackage{amsmath}
\usepackage{amssymb}

\usepackage[pagebackref=true,breaklinks=true,letterpaper=true,colorlinks,bookmarks=false]{hyperref}

\cvprfinalcopy 

\def\cvprPaperID{****} 

\ifcvprfinal\fi
\begin{document}

\title{Rethinking Continual Learning for Autonomous Agents and Robots}


\author{German I. Parisi\\
Apprente, Inc., Mountain View, CA\\
University of Hamburg, Germany\\
ContinualAI\\
{\tt\small parisi@informatik.uni-hamburg.de}
\and
Christopher Kanan\\
PaigeAI, New York, NY\\
Rochester Institute of Technology, Rochester, NY\\
Cornell Tech, New York, NY\\
{\tt\small kanan@rit.edu}
}

\maketitle



\section*{Introduction}

Continual learning refers to the ability of a biological or artificial system to seamlessly learn from continuous streams of information while preventing \textit{catastrophic forgetting}, i.e., a condition in which new incoming information strongly interferes with previously learned representations~\cite{Mermillod2013, Parisi2019}.
In the context of machine learning, continual learning models aim to reflect a number of properties of biological systems and their ability to acquire, fine-tune, and transfer knowledge and skills throughout a lifespan.
However, despite significant advances in continual machine learning, state-of-the-art models are still far from providing the flexibility, robustness, and scalability exhibited by biological systems.
The complexity of the datasets used for the evaluation of continual learning tasks is very limited and does not reflect the richness and level of uncertainty of the stimuli that artificial agents can be exposed to in the real world~(see~\cite{Parisi2019, Chen2018, Kemker2018a} for recent reviews).
Furthermore, neural models are often trained with data samples shown in isolation or presented in a random order.
This significantly differs from the highly organized manner in which humans and animals efficiently learn from samples presented in a meaningful order for the shaping of increasingly complex concepts and skills~\cite{Krueger2009}.
Therefore, learning in a continual manner goes beyond the incremental accumulation of domain-specific knowledge, enabling to transfer generalized knowledge and skills across multiple tasks and domains~\cite{Barnett2002} and, importantly, benefiting from the interplay of multisensory information for the development and specialization of complex neurocognitive functions~\cite{Murray2016}.

Since it is unrealistic to provide artificial agents with all the necessary prior knowledge to effectively operate in real-world conditions, they must exhibit a rich set of learning capabilities enabling them to interact in complex environments with the aim to process and make sense of continuous streams of (often uncertain) information~\cite{Hassabis2017}.
While the vast majority of continual learning models are designed to alleviate catastrophic forgetting on simplified classification tasks, here we focus on continual learning for autonomous agents and robots required to operate in much more challenging experimental settings.
In particular, we discuss well-established biological learning factors such as developmental and curriculum learning, transfer learning, and intrinsic motivation and their computational counterparts for modeling the progressive acquisition of increasingly complex knowledge and skills in a continual fashion.

\begin{figure*}[t]
\centering
\includegraphics[width=\textwidth]{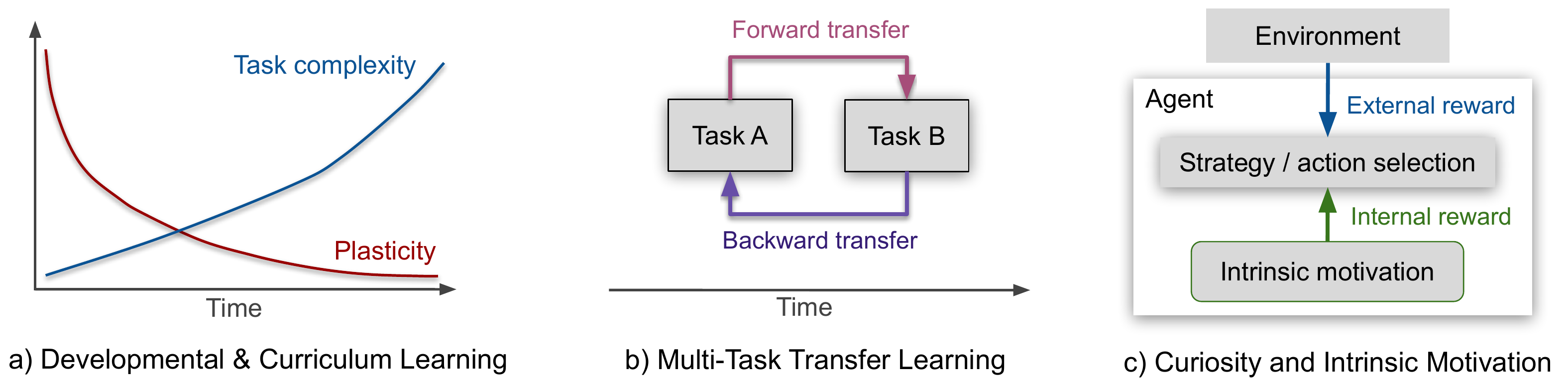}
	\caption{Schematic view of the main components for the development of continual learning autonomous agents. Adapted with permission from~\cite{Parisi2019}.}
\label{fig:learning}
\end{figure*}

\section*{Developmental and Curriculum Learning}

Humans and high animals show an exceptional capacity to learn throughout their lifespan and, with respect to other species, exhibit the lengthiest developmental process for reaching maturity.
There is a limited time window in development in which infants are particularly sensitive to the effects of their experiences.
This period is commonly referred to as \textit{critical} period of development in which early experiences are particularly influential~\cite{Senghas2004}.
During these critical periods, the brain is particularly plastic and neural networks acquire their overarching structure driven by sensorimotor experiences (see \cite{Power2016} for a survey; Fig.~\ref{fig:learning}.a).
Afterwards, plasticity becomes less prominent and the system stabilizes, preserving a certain degree of plasticity for subsequent reorganisation at smaller scales~\cite{Quadrato2014}.

Developmental learning strategies have been proposed to regulate the embodied interaction with the environment in real time~\cite{Cangelosi2015, Tani2016}.
In contrast to computational models that are fed with batches of information, developmental agents acquire an increasingly complex set of skills based on their sensorimotor experiences in an autonomous manner.
Consequently, staged development becomes essential for bootstrapping cognitive skills with less amount of tutoring experience.
However, the use of developmental strategies for artificial learning systems has shown to be a very complex practice.
In particular, it is difficult to select a well-defined set of developmental stages that favours the overall learning performance in highly dynamic environments.

Better learning performance is exhibited when examples are organized in a meaningful way, e.g., by making the learning tasks gradually more difficult~\cite{Krueger2009}.
It has been shown that having a \textit{curriculum} of progressively harder tasks leads to faster training performance in neural network systems~\cite{Elman1993}.
This has inspired similar approaches in robotics and more recent machine learning methods studying the effects of curriculum learning in the performance of learning~\cite{Graves2016}.
The task selection problem can be treated as a stochastic policy over the tasks that maximizes the learning progress, leading to an improved efficiency in curriculum learning~\cite{Graves2017}.
In this case, it is necessary to introduce additional factors such as intrinsic motivation~\cite{Barto2013}, where indicators of learning progress are used as reward signals to encourage exploration.
Curriculum strategies can be seen as a special case of transfer learning~\cite{Weiss2016}, where the knowledge collected during the initial tasks is used to guide the learning process of more sophisticated ones.

\section*{Transfer Learning}

Transfer learning applies previously acquired knowledge in one domain to solve a problem in a novel domain~\cite{Barnett2002}.
In this context, forward transfer refers to the influence that learning a task $\mathcal{T}_A$ has on the performance of a future task $\mathcal{T}_B$, whereas backward transfer refers to the influence of a current task $\mathcal{T}_B$ on a previous task $\mathcal{T}_A$~(Fig.~\ref{fig:learning}.b).
For this reason, transfer learning represents a significantly valuable feature of artificial systems for inferring general laws from (a limited amount of) particular samples, assuming the simultaneous availability of multiple learning tasks with the aim to improve the performance at one specific task.

Transfer learning has remained an open challenge in machine learning and autonomous agents (see \cite{Weiss2016} for a survey).
Specific neural mechanisms in the brain mediating the high-level transfer learning are poorly understood, although it has been argued that the transfer of abstract knowledge may be achieved through the use of conceptual representations that encode relational information invariant to individuals, objects, or scene elements~\cite{Doumas2008}.
Zero-shot learning~\cite{Lampert2009} and one-shot learning~\cite{FeiFei2003} aim at performing well on novel tasks but do not prevent catastrophic forgetting on previously learned tasks.
A recent model called the Gradient Episodic Memory \cite{LopezPaz2017} alleviates catastrophic forgetting and performs \textit{positive} transfer to previously learned tasks.
The model learns the subset of correlations common to a set of distributions or tasks, able to predict target values associated with previous or novel tasks without making use of task descriptors.

\section*{Curiosity and Intrinsic Motivation}

Computational models of intrinsic motivation have taken inspiration from the way human infants choose their goals and progressively acquire skills to define developmental structures in continual learning frameworks~(see \cite{Gottlieb2013} for a review; Fig.~\ref{fig:learning}.c).
Infants select experiences that maximize an intrinsic learning reward through an empirical process of exploration.
The intrinsically motivated exploration of the environment, e.g., driven by the maximization of the learning progress~\cite{Schmidhuber1991}, can lead to the self-organization of human-like developmental structures where the skills being acquired become progressively more complex.

Computational models of intrinsic motivation collect data and acquire skills incrementally through the online (self-)generation of a learning curriculum~\cite{Forestier2016}.
This allows the stochastic selection of tasks to be learned with an active control of the growth of the complexity.
Recent work in reinforcement learning has included mechanisms of curiosity and intrinsic motivation to address scenarios where the rewards are sparse or deceptive~\cite{Forestier2017}.
In a scenario with very sparse extrinsic rewards, curiosity-driven exploration provides intrinsic reward signals that enable the agent to autonomously learn increasingly complex tasks.

\section*{Conclusion and Open Challenges}

Continual learning represents a crucial but challenging component of artificial learning systems and autonomous agents operating on real-world data, which is typically non-stationary and temporally correlated.
Additional research efforts are required to combine multiple methodologies that integrate a variety of factors observed in human learners for the development of agents and robots learning in an autonomous fashion.

Basic mechanisms of critical periods of development can be modeled to empirically determine convenient multilayered neural network architectures and initial patterns of connectivity that improve the performance of the model for subsequent learning tasks.
Methods comprising curriculum and transfer learning are a fundamental feature for reusing previously acquired knowledge and skills to solve a problem in a novel domain by sharing low- and high-level representations.
Approaches using intrinsic motivation are crucial for the self-generation of goals, leading to an empirical process of exploration and the progressive acquisition of increasingly complex skills in a continual fashion.


{\small
\bibliographystyle{ieee}
\bibliography{mybibNN}

\begin{thebibliography}{10}\itemsep=-1pt

\bibitem{Barnett2002}
S.~Barnett and S.~Ceci.
\newblock When and where do we apply what we learn? a taxonomy for far
  transfer.
\newblock {\em Psychological Bulletin}, 128:612--637, 2002.

\bibitem{Barto2013}
A.~Barto.
\newblock {\em Intrinsic motivation and reinforcement learning}.
\newblock Baldassarre, G., Mirolli, M. (Eds.), Intrinsically Motivated Learning
  in Natural and Artificial Systems. Springer, 2013.

\bibitem{Cangelosi2015}
A.~Cangelosi and M.~Schlesinger.
\newblock {\em Developmental robotics: From babies to robots}.
\newblock MIT Press, 2015.

\bibitem{Chen2018}
Z.~Chen and B.~Liu.
\newblock Lifelong machine learning: Second edition.
\newblock Morgan \& Claypool Publishers, 2018.

\bibitem{Doumas2008}
L.~Doumas, J.~Hummel, and C.~Sandhofer.
\newblock A theory of the discovery and predication of relational concepts.
\newblock {\em Psychological Review}, 115:1--43, 2008.

\bibitem{Elman1993}
J.~L. Elman.
\newblock Learning and development in neural networks: The importance of
  starting small.
\newblock {\em Cognition}, 48(1):71--99, 1993.

\bibitem{FeiFei2003}
L.~Fei-Fei, R.~Fergus, and P.~Perona.
\newblock A bayesian approach to unsupervised one-shot learning of object
  categories.
\newblock ICCV'03, Nice, France, 2003.

\bibitem{Forestier2017}
S.~Forestier, Y.~Mollard, and P.-Y. Oudeyer.
\newblock Intrinsically motivated goal exploration processes with automatic
  curriculum learning.
\newblock arXiv:1708.02190, 2017.

\bibitem{Forestier2016}
S.~Forestier and P.-Y. Oudeyer.
\newblock Curiosity-driven development of tool use precursors: a computational
  model.
\newblock Proceedings of the Annual Conference of the Cognitive Science
  Society, 2016.

\bibitem{Gottlieb2013}
J.~Gottlieb, P.-Y. Oudeyer, M.~Lopes, and A.~Baranes.
\newblock Information seeking, curiosity and attention: Computational and
  neural mechanisms.
\newblock {\em Trends in Cognitive Science}, 17(11):585--596, 2013.

\bibitem{Graves2017}
A.~Graves, M.~G. Bellemare, J.~Menick, R.~Munos, and K.~Kavukcuoglu.
\newblock Automated curriculum learning for neural networks.
\newblock arXiv:1704.03003, 2017.

\bibitem{Graves2016}
A.~Graves, G.~Wayne, M.~Reynolds, T.~Harley, I.~Danihelka,
  A.~Grabska-Barwinska, S.~G. Colmenarejo, E.~Grefenstette, T.~Ramalho, and
  J.~e.~a. Agapiou.
\newblock Hybrid computing using a neural network with dynamic external memory.
\newblock {\em Nature}, 538:471--476, 2016.

\bibitem{Hassabis2017}
D.~Hassabis, D.~Kumaran, C.~Summerfield, and M.~Botvinick.
\newblock Neuroscience-inspired artificial intelligence.
\newblock {\em Neuron Review}, 95(2):245--258, 2017.

\bibitem{Kemker2018a}
R.~Kemker, M.~McClure, A.~Abitino, T.~Hayes, and C.~Kanan.
\newblock Measuring catastrophic forgetting in neural networks.
\newblock AAAI'18, New Orleans, LA, 2018.

\bibitem{Krueger2009}
K.~A. Krueger and P.~Dayan.
\newblock Flexible shaping: how learning in small steps helps.
\newblock {\em Cognition}, 110:380--394, 2009.

\bibitem{Lampert2009}
C.~Lampert, H.~Nickisch, and S.~Harmeling.
\newblock Learning to detect unseen object classes by between-class attribute
  transfer.
\newblock CVPR'09, Miami Beach, Florida, 2009.

\bibitem{LopezPaz2017}
D.~Lopez-Paz and M.~Ranzato.
\newblock Gradient episodic memory for continual learning.
\newblock NIPS'17, Long Beach, CA, 2017.

\bibitem{Mermillod2013}
M.~Mermillod, A.~Bugaiska, and P.~Bonin.
\newblock The stability-plasticity dilemma: Investigating the continuum from
  catastrophic forgetting to age-limited learning effects.
\newblock {\em Frontiers in Psychology}, 4(504), 2013.

\bibitem{Murray2016}
M.~M. Murray, D.~J. Lewkowicz, A.~Amedi, and M.~T. Wallace.
\newblock Multisensory processes: A balancing act across the lifespan.
\newblock {\em Trends in Neurosciences}, 39:567--579, 2016.

\bibitem{Parisi2019}
G.~I. Parisi, R.~Kemker, J.~L. Part, C.~Kanan, and S.~Wermter.
\newblock Continual lifelong learning with neural networks: A review.
\newblock {\em Neural Networks}, 113:54--71, 2019.

\bibitem{Power2016}
J.~D. Power and B.~L. Schlaggar.
\newblock Neural plasticity across the lifespan.
\newblock {\em Wiley Interdisciplinary Reviews: Developmental Biology}, 6(216),
  2016.

\bibitem{Quadrato2014}
G.~Quadrato, M.~Y. Elnaggar, and S.~Di~Giovanni.
\newblock Adult neurogenesis in brain repair: Cellular plasticity vs. cellular
  replacement.
\newblock {\em Frontiers in Neuroscience}, 8(17), 2014.

\bibitem{Schmidhuber1991}
J.~Schmidhuber.
\newblock Curious model-building control systems.
\newblock 1991.

\bibitem{Senghas2004}
A.~Senghas, S.~Kita, and A.~{\"O}zy{\"u}rek.
\newblock Children creating core properties of language: Evidence from an
  emerging sign language in {Nicaragua}.
\newblock {\em Science}, 305:1779--1782, 2004.

\bibitem{Tani2016}
J.~Tani.
\newblock {\em Exploring Robotic Minds: Actions, Symbols, and Consciousness a
  Self-Organizing Dynamic Phenomena}.
\newblock Oxford University Press, 2016.

\bibitem{Weiss2016}
K.~Weiss, T.~M. Khoshgoftaar, and D.-D. Wang.
\newblock A survey of transfer learning.
\newblock {\em Journal of Big Data}, 3(9), 2016.

\end{thebibliography}
}

\end{document}